\documentclass[10pt]{article}
\usepackage[preprint]{tmlr}
\usepackage[table]{xcolor}
\usepackage{enumitem}
\usepackage{latexsym}
\usepackage{wrapfig}
\usepackage{booktabs}
\usepackage{array}
\usepackage{longtable}
\usepackage{comment}
\usepackage{booktabs}
\usepackage{tabularx}
\usepackage[T1]{fontenc}
\usepackage{amsmath}

\definecolor{lightpurple}{RGB}{242, 235, 250}
\definecolor{lightblue}{RGB}{232, 242, 252}
\definecolor{lightgreen}{RGB}{235, 248, 239}
\definecolor{lightgray}{RGB}{245,245,245}

\usepackage[utf8]{inputenc}

\usepackage{microtype}

\usepackage{inconsolata}

\usepackage{graphicx}
\graphicspath{{../}{./}}
\usepackage{tcolorbox}
\usepackage{hyperref}
\usepackage{url}
\usepackage{subcaption}
\title{
\includegraphics[width=0.15\linewidth]{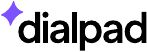}\\[0.5em]
From Text to Voice: A Reproducible and Verifiable \\ Framework for Evaluating Tool Calling LLM Agents}

\author{\name Md Tahmid Rahman Laskar \email tahmid.rahman@dialpad.com\\
      \addr Dialpad Inc.
      \AND
      \name Xue-Yong Fu \email xue-yong@dialpad.com\\
      \addr Dialpad Inc.
      \AND
      \name Seyyed Saeed Sarfjoo \email saeed.sarfjoo@dialpad.com\\
      \addr Dialpad Inc.
      \AND
      \name Quinten McNamara \email quinn.mcnamara@dialpad.com\\
      \addr Dialpad Inc.
      \AND
      \name Jonas Robertson \email jonas@dialpad.com\\
      \addr Dialpad Inc.
      \AND
      \name Shashi Bhushan TN \email sbhushan@dialpad.com\\
      \addr Dialpad Inc.}

\def\month{MM}
\def\year{YYYY}
\def\openreview{\url{https://openreview.net/forum?id=XXXX}}

\begin{document}
\maketitle
\begin{abstract}

Voice agents increasingly require reliable tool use from speech, whereas prominent tool-calling benchmarks remain text-based. We study whether verified text benchmarks can be converted into controlled audio-based tool calling evaluations without re-annotating the tool schema and gold labels. Our dataset-agnostic framework uses text-to-speech, speaker variation, and environmental noise to create paired text--audio instances while preserving the original dataset annotations. 
Based on extensive evaluation of 7 omni-modal models on audio-converted versions of Confetti and When2Call, 
our framework demonstrates that the performance is strongly model- and task-dependent: Gemini-3.1-Flash-Live obtains the highest Confetti score (70.4), whereas GPT-Realtime-1.5 performs best on When2Call (71.9). On Confetti, the text-to-voice gap ranges from 1.8 points for Qwen3-Omni to 4.8 points for GPT-Realtime-1.5. A targeted analysis of failure cases demonstrates that degradations most often reflect misunderstandings of argument values in the speech. 
Considering real-world deployment scenarios, we further report text-only 
results, an ambiguity-based reformulation stress test, and a reference-free LLM-as-judge protocol validated against human preferences. Notably, we find that open-source Qwen3 judges with at least 8B parameters exceed 80\% agreement with proprietary judges, supporting privacy-preserving evaluation. Overall, our framework provides a verifiable and reproducible first-stage diagnostic that complements purpose-built audio corpora.

\end{abstract}

\section{Introduction}
The rapid advancement of Large Language Models (LLMs) has enabled sophisticated tool calling and function execution capabilities \citep{schick2023toolformer,qin2023toolllm,qin2024tool,qu2025tool}, in which models identify when to use external functions, determine appropriate parameters, and integrate results into coherent responses. Tool calling has become a cornerstone of modern Agentic AI systems with benchmarks like Confetti \citep{alkhouli-etal-2025-confetti}, When2Call \citep{ross-etal-2025-when2call}, and BFCL \citep{patil2025bfcl}, 
have already absorbed the substantial cost of validating tool schemas, curating gold labels, and designing scoring protocols.

Yet in customer support, where interactions often occur via voice \citep{rzepka2022voice,laskar2025ai}, these text benchmarks cannot be applied directly. Practitioners deploying voice agents face a concrete architectural choice: a \textit{cascade} pipeline that runs an Asynchronous Speech Recognition (ASR) system followed by a text LLM \citep{peng2025survey}, or an \textit{end-to-end omni-modal LLM} that processes speech directly and emits tool calls without an intermediate transcript \citep{jiang2025specific}. 
Cascades usually leverage strong text-only LLMs, expose transcripts for debugging, and make failures easier to localize across ASR, tool selection, and argument generation \citep{lin2025neko}. However, they can propagate ASR errors, lose acoustic cues, and increase latency \citep{yang2023taskactivating}. End-to-end omni-modal models avoid the transcription bottleneck, but their tool-calling reliability under realistic audio conditions remains insufficiently characterized.

Existing audio function-calling benchmarks \citep{bfclaudio2025,voiceagentbench2025} 
mainly follow a dataset-construction pattern: build a new audio corpus, define new tools, annotate new labels, and report a leaderboard. Although useful, such benchmarks are often restricted to the domains and tool schemas chosen during construction. This is less helpful for organizations that already have validated domain-specific text benchmarks and need to know how those tasks behave when deployed over voice. 

To this end, we study whether existing text-based tool-calling benchmarks can be systematically converted into speech benchmarks while preserving their gold labels and tool schemas. This enables paired text--audio evaluation, measures the degradation introduced by speech, and provides an initial diagnostic for comparing the cascade architecture with end-to-end omni-modal models. More specifically, we introduce a dataset-agnostic framework that converts any text-based tool-calling benchmark into a controlled audio evaluation using off-the-shelf TTS models \citep{tan2021survey}, while preserving the original tool schemas, gold labels, and evaluation protocols. Rather than proposing another standalone audio benchmark, our contribution is a reusable recipe that enterprises can apply to proprietary text-based tool-calling data. 
We treat TTS-generated speech as a controlled 
approximation of voice-based interaction, 
which enables us to isolate text--audio and cascade--omni performance gaps. Compared with building audio benchmarks from scratch, this design has three advantages: it preserves gold labels and tool schema, creates paired text--audio instances for failure analysis, and can be applied to proprietary enterprise tool catalogs and text logs. 

Using this framework, we evaluate seven omni-modal models across four providers and multiple model tiers, including GPT-Realtime\footnote{\url{https://developers.openai.com/api/docs/guides/realtime}} (4o, Mini, 1.5), Gemini-Flash-Live\footnote{\url{https://ai.google.dev/gemini-api/docs/live-api}} (2.5 and 3.1), Qwen3-Omni \citep{Qwen3-Omni}, and Phi-4-Multimodal \citep{abouelenin2025phi}. We compare these systems with 
text-only baselines on \textit{Confetti} and \textit{When2Call} under clean and noisy audio with diverse voices. Since the cascade architecture depends on the strength of the downstream text LLM, we further conduct a text-mode scaling analysis of Qwen3 models from 0.6B to 32B parameters. We also include an ambiguity-based query reformulation stress test and an error analysis of the text-to-voice performance gap. Finally, because production data often lacks gold annotations, we assess reference-free LLM-as-judge protocols for tool-calling evaluation using both proprietary and open judges \citep{gu2024surveyllmjudge}. Our results show that converting trusted text benchmarks gives a signal on architectural choices and failure modes that fixed audio corpora may not isolate directly. We argue that audio-converted versions of trusted text benchmarks can complement purpose-built audio datasets. 

Our major contributions are summarized below:

\begin{enumerate}[label=\roman*., itemsep=-3pt, topsep=2pt]

\item \textbf{Benchmark conversion for verifiable audio tool-calling evaluation.} We introduce a dataset-agnostic framework for converting existing text-based tool-calling benchmarks into controlled audio evaluations using TTS, voice variation, and environmental noise. Unlike fully new audio benchmarks, this approach preserves the original tool schemas, gold labels, and verifiable scoring protocol. 

\item \textbf{A diagnostic protocol for paired text--audio comparison.}
By evaluating matched text and audio versions of the same tool-calling instances, our framework measures voice-input degradation relative to a known text baseline and provides a practical signal for modality-induced failures.

\item \textbf{Model-level evidence for cascade-vs-omni diagnostics.}
We show that audio tool-calling performance is model- and task-dependent, not determined by architecture alone. Model rankings change across benchmarks, and the Confetti text-to-voice gap ranges from 1.8 to 4.8 points across various models, motivating model- and task-specific cascade-vs-omni diagnostics.

\item \textbf{An error decomposition enabled by paired benchmark conversion.}
Because every audio instance has a paired text version with the same gold label, we isolate cases where a model succeeds on text but fails on the corresponding audio input. This enables a counterfactual error analysis showing that audio-induced failures often preserve the broad tool-call structure but fail on argument values, especially for Gemini-3.1-Flash-Live and GPT-Realtime-1.5.

\item \textbf{A reference-free LLM-as-judge protocol for production-style evaluation.}
Since production data often lacks gold annotations, we assess reference-free LLM-as-judge protocols using both proprietary and open judges. Proprietary judges show broadly stable model-level rankings, and a human preference study 
finds no clear preference between GPT-5 and Gemini-2.5-Pro. We also observe that open-source Qwen3 judges with at least 8B parameters reach over 80\% agreement with proprietary judges, suggesting a path toward privacy-preserving evaluation with 
open judges.


\end{enumerate}

As a secondary contribution, our converted datasets and evaluation scripts are made publicly available here: \url{https://github.com/talkiq/dialpad-ai-research/tree/main/toolvoice}. 
\section{Related Work}
\noindent \textbf{Tool Calling using LLMs:}
Tool calling extends LLM capabilities by enabling interaction with external APIs and computational tools \citep{mialon2023augmented}. Prior work has explored structured API induction via fine-tuning \citep{li2023apibank,xu2023tool,schick2023toolformer,patil2024gorilla} and introduced large-scale benchmarks such as ToolBench and API-Bank \citep{qin2023toolllm,li2023apibank}, along with synthetic scaling approaches like ToolAlpaca \citep{tang2023toolalpaca}. Domain-specific benchmarks target areas such as mathematics \citep{gou2023tora} and customer support \citep{alkhouli-etal-2025-confetti,ross-etal-2025-when2call}, while $\tau$-Bench provides unified agentic evaluation \citep{yao2025tau}. Crucially, these benchmarks have already absorbed high cost in validating tool specifications, curating gold parameter values, and designing scoring protocols. However, they remain primarily text-only and cannot directly evaluate omni-modal models that operate on speech. Our framework closes this gap by transforming existing text-based tool-calling benchmarks into audio evaluations while preserving their validated tool schemas, gold labels, and scoring protocols. 

\noindent \textbf{Omni-Modal Large Language Models:}
Omni-modal LLMs extend text-only systems to unified reasoning across text, vision, and audio \citep{jiang2025specific}. Unlike cascade architectures that rely on speech-to-text pipelines, end-to-end omni-modal models process audio natively and can reduce pipeline complexity. Both proprietary (e.g., GPT-4o-Realtime, Gemini-Live) and open-source models \citep{yao2024minicpm,zeng2024glm4,Qwen2.5-Omni,kimiteam2025kimiaudiotechnicalreport,Qwen3-Omni} support such capabilities. Nevertheless, their reliability for structured tool calling remains unsettled, which we aim to address using our proposed framework. 

\noindent \textbf{Audio Understanding Benchmarks:}
Most of the existing audio benchmarks focus on perception tasks such as ASR \citep{librispeech2015,ardila2020common}, captioning  \citep{kim2019audiocaps,drossos2020clotho}, and event classification \citep{gemmeke2017audioset,huang2024dynamic}. More recent benchmarks (e.g., VoiceBench, WildSpeech-Bench) assess spoken interaction robustness \citep{chen2024voicebench,zhang2025wildspeech}. These resources are important for speech robustness, but they do not provide the 
tool-calling evaluation setup, which is the focus of this work. 

\noindent \textbf{Tool Calling Audio Benchmarks:}
Evaluation is critical for LLM deployment \citep{laskar2023systematicchatgpt,laskar2024systematic}, and recent work has begun evaluating tool calling from speech, including BFCL-v4's audio tier \citep{patil2025bfcl,bfclaudio2025}, as well as benchmarks like VoiceAgentBench \citep{voiceagentbench2025} 
These benchmarks construct new audio corpora in selected domains and report the model performance. 
Our work is complementary: instead of constructing another fixed audio corpus, we convert existing text-based tool-calling benchmarks into paired speech evaluations. This design supports deployment-oriented analysis in three ways. First, it enables system-level comparison: practitioners can evaluate specific ASR$\rightarrow$text-LLM cascade pipelines and omni-modal models on the same tasks, rather than assuming one architecture is always better. Second, each audio example has a clean-text counterpart with the same gold label, allowing direct measurement of the text-to-audio gap and instance-level error decomposition. Third, the framework is portable to proprietary enterprise tool catalogs and text logs, enabling in-house audio evaluation on annotated datasets. 
Therefore, 
existing benchmarks answer the question ``how do current SpeechLMs perform on a fixed audio corpus for tool calling?'', while we answer ``how should an enterprise team decide between cascade and omni on its own data for agentic tasks, and what does the modality shift actually cost?''
\begin{figure*}
    \centering
    \includegraphics[width=\linewidth]{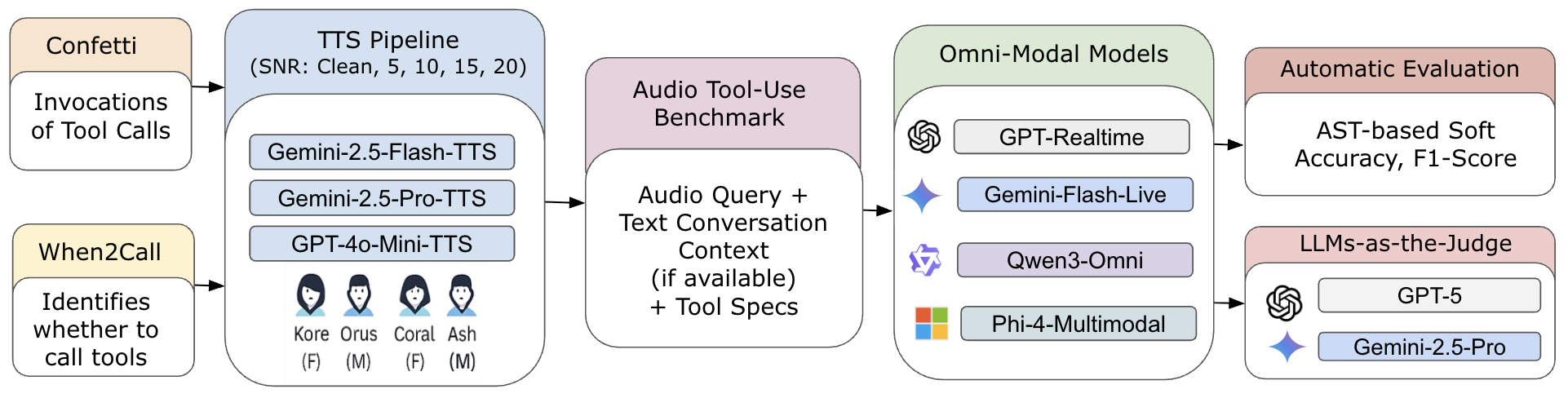}
    \caption{\small{An overview of our methodology for converting text-based tool datasets into audio benchmarks for tool-calling evaluation. The pipeline uses text-to-speech (TTS) models (GPT-4o-Mini-TTS and Gemini-2.5-TTS) to generate diverse audio queries with different voices and genders, which are then processed by omni-modal LLMs and evaluated 
    via automatic evaluation or LLM Judge.}}
    \label{fig:overview}
\end{figure*}
\section{Methodology}
\subsection{Text-to-Speech Conversion Pipeline}
We implement a systematic TTS conversion pipeline with three commercially available TTS models that can be accessed through API endpoints:

\noindent \textbf{(i) Gemini-2.5-Flash-TTS\footnote{\url{https://docs.cloud.google.com/text-to-speech/docs/gemini-tts\#gemini-2-5-flash-tts}}}: Google's efficient, low-latency TTS model. 
 
\noindent \textbf{(ii) Gemini-2.5-Pro-TTS\footnote{\url{https://docs.cloud.google.com/text-to-speech/docs/gemini-tts\#gemini-2-5-pro-tts}}}: A higher-capacity Gemini TTS system optimized for speech generation quality.

\noindent \textbf{(iii) GPT-4o-Mini-TTS\footnote{\url{https://platform.openai.com/docs/models/GPT-4o-Mini-TTS}}}: A compact OpenAI TTS model used as a cross-provider alternative.

By comparing Gemini-2.5-Flash-TTS with the higher-capacity Gemini-2.5-Pro-TTS, we examine whether increased model scale and associated cost translate into measurable gains in synthesis quality and robustness under diverse noisy environments. Additionally, GPT-4o-Mini-TTS serves as a cross-provider alternative to Gemini-2.5-Flash-TTS, enabling us to examine whether compact models from different providers exhibit performance variation under comparable efficiency constraints.

\textbf{Speaker Variation}: To assess model robustness across speaker characteristics, we select distinct voices from each TTS provider for both genders. For Gemini voices, \textit{Kore} represents female voice, while  \textit{Orus} represents male. For {GPT-4o-Mini}, \textit{Ash} represents male voice and \textit{Coral} represents female. 

\textbf{Environmental Noise Injection}:
To simulate controlled deployment-like acoustic conditions, we inject environmental noise from the DEMAND dataset~\citep{thiemann2013demand}, which contains multi-channel recordings of acoustic noise across diverse environments. We sample noise segments from settings such as cars, buses, traffic, cafés, kitchens, hallways, meeting rooms, and living rooms, and mix them with clean speech. To evaluate robustness under varying noise levels, we construct separate subsets by adding noise at 5, 10, 15, and 20 dB signal-to-noise ratios (SNRs). Higher SNR values correspond to cleaner audio, reflecting a larger proportion of clean signal energy relative to noise.

\subsection{TTS Audio as a Controlled First-Stage Evaluation}
Our benchmark is designed as an early-stage evaluation tool for companies making decisions about deploying voice-enabled LLM agents. This is especially useful for teams that already have annotated text transcripts and tool logs: our TTS conversion pipeline provides a low-friction way to test whether a candidate voice-agent stack is likely to preserve tool-calling behavior when the input changes from text to speech. We therefore treat TTS audio as an \emph{optimistic deployment proxy} and first-stage evaluation strategy, rather than a replacement for spontaneous real calls.
The key design choice is to evaluate each task in two aligned conditions. In the \emph{text-only} condition, the original text input is given directly to the LLM, measuring the model's downstream ability to use the correct tool 
when speech processing is not involved. In the \emph{direct-audio} condition, the TTS version of the same input is given to an omni-modal model, measuring whether the model can perform the same tool-calling task from speech. Comparing the text and audio versions of the same examples allows us to estimate the performance loss caused by changing the input modality from text to speech while keeping the task, tool schema, and gold labels fixed.
This comparison thereby provides a practical signal for whether an existing text benchmark can support controlled audio evaluation and offers an initial indication of how direct omni-modal inference compares with an ASR$\rightarrow$text cascade.. The claim is intentionally conservative: if controlled TTS audio already exposes substantial failures, spontaneous real-world speech is likely to pose an even harder setting.
 \subsection{Preserving Verifiability through Benchmark Conversion}
Building an audio tool-calling benchmark from scratch is costly because it requires defining tool schemas, annotating the correct tool calls and arguments, deciding when a tool call is required, and designing reliable evaluation protocols. These steps are especially challenging for speech input, where evaluation must account for acoustic variation, speaker differences, and possible transcription variability in cascade systems \citep{sofer2025pull}. At the same time, text-based tool-calling resources are often more readily available and have already absorbed much of this cost. For instance, public benchmarks such as ToolBench, Confetti, When2Call, and BFCL already provide validated tool specifications, gold parameter values, decision labels, and established scorers \citep{qin2023toolllm,alkhouli-etal-2025-confetti,ross-etal-2025-when2call,patil2025bfcl}; many enterprise teams similarly maintain text transcripts, tool logs, API schemas, and text-based evaluation pipelines. Our conversion pipeline reuses these resources rather than rebuilding them from scratch.

This design has two practical benefits. First, it preserves verifiability: the converted audio benchmark inherits the source benchmark's tool schemas, gold labels, and evaluation protocol. Second, it enables a controlled text-to-speech comparison: each audio input is generated from an existing text instance, allowing us to measure how performance changes when the same task is presented as speech. The conversion is not merely a cheaper data-generation strategy; it preserves the source benchmark’s tool schemas, gold labels, and scoring protocol while testing the effect of changing only the input modality. As the source benchmark improves through better labels, harder splits, or broader tool coverage, the converted audio benchmark can improve with it. We therefore view audio conversion as complementary to purpose-built audio datasets. Speech-native corpora are valuable for studying naturally spoken interactions, while conversion is useful for testing whether existing domain-specific text-based tool-calling tasks remain reliable under voice input. 


\subsection{Datasets}

We employ two publicly available text-based tool-calling benchmarks that are representative of customer-support and conversational agent scenarios for our audio conversion pipeline. While we select two datasets in this paper, our audio construction pipeline is dataset-agnostic: any text-based tool-calling benchmark can be integrated into our framework (see Figure \ref{fig:overview}) without any architectural modification. 

\noindent \textbf{(i) Confetti} \citep{alkhouli-etal-2025-confetti}: Confetti is a high-quality tool-calling dataset covering diverse real-world conversational settings. Each instance in this dataset includes: (i) a natural language user query requiring tool invocation, (ii) a multi-turn conversational context, and (iii) tool/API documentation with function signatures and parameter specifications. Since our evaluation focuses on tool execution correctness, we consider only those instances that require an explicit tool call, resulting in a filtered subset of 313 examples.

\noindent \textbf{(ii) When2Call} \citep{ross-etal-2025-when2call}: This dataset evaluates the model’s ability to determine whether a tool call is necessary, as opposed to directly responding to the user or requesting clarification. Each instance consists of: (i) a user utterance that may or may not require tool invocation, and (ii) the corresponding API/tool specifications. We use the non-MCQ subset of the dataset, which contains 300 instances.

These benchmarks provide diverse evaluation perspectives. Confetti measures function selection and parameter extraction accuracy, whereas When2Call assesses the decision-making capability required to determine whether to call a tool or not. 

\subsection{Omni-Modal Models}

We primarily evaluate seven 
omni-modal models from four providers, spanning multiple model tiers. 

\noindent \textbf{GPT-4o-Realtime}: GPT-4o-Realtime\footnote{\url{https://developers.openai.com/api/docs/models/gpt-4o-realtime-preview}} is designed for real-time voice-to-voice interaction. It processes audio input natively and generates streaming speech responses alongside direct tool calling from voice input.

\noindent \textbf{GPT-Realtime-1.5}: This Realtime\footnote{\url{https://developers.openai.com/api/docs/models/gpt-realtime-1.5}} model is an updated version of GPT-4o-Realtime. We evaluate it through the same streaming interface that we used for GPT-4o-Realtime. 

\noindent \textbf{GPT-Realtime-Mini}: We evaluate this compact OpenAI Realtime\footnote{\url{https://developers.openai.com/api/docs/models/gpt-realtime-mini}} variant to measure whether a lower-cost model condition preserves audio tool-calling behavior.

\noindent \textbf{Gemini-2.5-Flash-Live}: We use the Gemini-2.5-Flash-Native-Audio model via Google's Live API\footnote{\url{https://docs.cloud.google.com/vertex-ai/generative-ai/docs/models/gemini/2-5-flash-live-api}}. It is a real-time multimodal model optimized for low-latency interactive applications, supporting native audio input, streaming speech generation, and structured function calling.

\noindent \textbf{Gemini-3.1-Flash-Live}: This is the latest omni-modal model from Google\footnote{\url{https://ai.google.dev/gemini-api/docs/models/gemini-3.1-flash-live-preview}}, which we evaluate through the same Live API interface we used for Gemini-2.5-Flash-Live.

\noindent \textbf{Qwen3-Omni-30B-A3B-Instruct}: Qwen3-Omni~\citep{Qwen3-Omni} is an end-to-end open omni-modal model built by leveraging the Thinker--Talker \citep{Qwen2.5-Omni} and the Mixture-of-Experts architecture \citep{shazeer2017outrageously}. The Thinker module handles reasoning and text generation, while the Talker module enables streaming speech generation. The model supports multimodal inputs (text, image, audio, video), streaming interaction with low latency, and function calling.

\noindent \textbf{Phi-4-Multimodal}: Microsoft's Phi-4-Multimodal-Instruct \citep{abouelenin2025phi} is a compact open-source omni-modal model (5.6B parameters) supporting text, image, and audio inputs. We include it as an additional compact open-source baseline for audio tool-calling behavior.

\section{Experiments}
In this section, we first demonstrate the implementation details, followed by the evaluation settings. Finally, we discuss the experimental results. 
\subsection{Implementation}
All audio inputs are stored in \texttt{.WAV} format and converted to 16\,kHz mono 16-bit PCM before streaming to model endpoints through WebSocket connections. We implement the OpenAI models (\texttt{GPT-4o-Realtime}, \texttt{GPT-Realtime-1.5}, and \texttt{GPT-Realtime-Mini}) via the OpenAI Realtime API using WebSocket streaming. The Gemini models (\texttt{Gemini-2.5-Flash-Live} and \texttt{Gemini-3.1-Flash-Live}) are implemented via the Google GenAI Live API\footnote{\url{https://ai.google.dev/gemini-api/docs/live-api}}. For \texttt{Qwen3-Omni}, we use the \textit{Qwen3-Omni-30B-A3B-Instruct} checkpoint and run inference using HuggingFace \citep{wolf2019huggingface}. For \texttt{Phi-4-Multimodal}, we also use the \textit{Phi-4-multimodal-instruct} checkpoint from HuggingFace and run inference using it. For all the text-only LLMs, we use the Qwen-3 series \citep{yang2025qwen3} models and run inference using vLLM \citep{vllm}. We use default decoding parameters for all models, except that temperature is set to 0.6 when the parameter is supported. The prompts used for response generation are provided in Appendix \ref{omni_llm_prompt}. 
\subsection{Evaluation Settings}
In When2Call, we assess whether the 
model response is a tool call or not by comparing the model-generated response against the ground truth using a parsing script \citep{laskar-etal-2024-query,saini2025llm} and report the F1-score. 

 In Confetti, we follow the standard evaluation protocol from the original paper \citep{alkhouli-etal-2025-confetti} to assess whether the model can accurately invoke the functions with the correct parameters. Specifically, we compare the model predicted function calls with the ground truth using an AST-based soft accuracy metric where the predicted function name and non-string parameter values are scored using exact match, while string parameter values are scored using AlignScore \citep{zha-etal-2023-alignscore}). 
Given the difficulty in evaluating the tool calling outputs generated by LLMs in real-world settings due to the unavailability of annotated reference labels, 
we also conduct a reference-free evaluation in Confetti using the following two LLMs-as-judge models \citep{zheng2023judging}: (i) GPT-5 \citep{openai2025gpt5systemcard}, (ii) Gemini-2.5-Pro \citep{comanici2025gemini}.


\begin{table}[t!]
\centering
\scriptsize
\setlength{\tabcolsep}{2.5pt}
\begin{tabular}{l|cc|cc|cc|cc|cc}
\toprule
& \multicolumn{2}{c}{\textbf{Clean}} & \multicolumn{2}{c}{\textbf{SNR 5}} & \multicolumn{2}{c}{\textbf{SNR 10}} & \multicolumn{2}{c}{\textbf{SNR 15}} & \multicolumn{2}{c}{\textbf{SNR 20}} \\
\cmidrule(lr){2-3} \cmidrule(lr){4-5} \cmidrule(lr){6-7} \cmidrule(lr){8-9} \cmidrule(lr){10-11}
\textbf{TTS Model} & \textbf{UTMOS} $\uparrow$ & \textbf{WER} $\downarrow$ & \textbf{UTMOS} $\uparrow$ & \textbf{WER} $\downarrow$ & \textbf{UTMOS} $\uparrow$ & \textbf{WER} $\downarrow$ & \textbf{UTMOS} $\uparrow$ & \textbf{WER} $\downarrow$ & \textbf{UTMOS} $\uparrow$ & \textbf{WER} $\downarrow$ \\
\midrule
\rowcolor{lightblue} Gemini-2.5-Flash & \underline{\textbf{3.51}} & \textbf{4.75} & 2.81 & 7.15 & 2.88 & 7.35 & 2.96 & 7.34 & 3.07 & 7.33 \\
\rowcolor{lightblue} Gemini-2.5-Pro   & \textbf{3.44} & \textbf{4.93} & 2.75 & 8.19 & 2.79 & 8.15 & 2.85 & 8.18 & 2.99 & 8.17 \\
\rowcolor{lightgray} GPT-4o-Mini      & \textbf{3.48} & \underline{\textbf{3.38}} & 2.84 & 4.34 & 2.87 & 4.17 & 2.92 & 4.25 & 3.00 & 4.19 \\
\bottomrule
\end{tabular}
\caption{\small TTS quality across the \textit{Confetti} and \textit{When2Call} datasets based on average across all voice types and genders. Higher UTMOS indicates better perceptual quality, while lower WER indicates better recognition accuracy. \underline{Bold} denotes the best scores. 
} 
\label{tab:tts-eval-dataset-avg}
\end{table}

\subsection{Results and Discussion}
We first report TTS quality evaluation results and then present omni-modal model performance on Confetti and When2Call datasets. 
\subsubsection{TTS Performance}
To evaluate synthetic speech quality, we use the UTokyo-SaruLab MOS Prediction System (UTMOSv2)~\citep{baba2024utmosv2}, which predicts the mean opinion score (MOS) for speech naturalness. Higher UTMOS scores indicate more natural synthetic speech. To assess intelligibility, we decode the synthetic audio using Whisper large-v3~\citep{radford2022whisper} and compute word error rate (WER), where lower WER indicates better intelligibility. We evaluate the TTS quality and report the results in Table~\ref{tab:tts-eval-dataset-avg}. 

Across clean and noisy conditions, GPT-4o-Mini obtains the lowest WER among the evaluated TTS models while maintaining competitive naturalness scores. Under clean conditions, Gemini-2.5-Flash-TTS obtains the highest UTMOS score, although the margin over GPT-4o-Mini is small. As expected, audio quality decreases under noise, with larger drops at lower SNR values. The relative ordering of TTS models is nevertheless stable across SNR levels, and GPT-4o-Mini maintains strong intelligibility under degraded audio conditions.

We use WER only as an audio-quality diagnostic and do not filter examples by ASR output; all generated samples are retained in the benchmark. To verify that the synthesized audio remains faithful to the source utterances beyond automatic metrics, we additionally conduct a human intelligibility check on 300 randomly sampled clips, split evenly between clean and noisy audio, using two domain experts having expertise in Natural Language Processing (NLP) and Speech Processing. Human evaluators mark 97.7\% of clean samples and 94.3\% of noisy samples as content-faithful, indicating that the benchmark largely preserves the intended query semantics while still exposing models to controlled audio degradation.

\subsubsection{Omni-Modal Model Performance}

\begin{table*}[t!]
\centering
\scriptsize
\setlength{\tabcolsep}{3pt}
\begin{tabular}{ll ccccccc}
\toprule
\textbf{TTS Model} & \textbf{Voice} &  \textbf{GPT-4o} & \textbf{GPT-1.5} & \textbf{GPT-Mini} & \textbf{Gemini-2.5} & \textbf{Gemini-3.1} & \textbf{Qwen3-Omni} &
\textbf{Phi-4}
\\
& & \textbf{Realtime} & \textbf{Realtime} & \textbf{Realtime} & \textbf{Flash-Live} & \textbf{Flash-Live} & \textbf{30B-A3B} &
\textbf{MM}
\\

\midrule
\rowcolor{lightblue}
 Gemini-2.5-Flash & Kore \textsc{(F)} & 51.0 & 59.4 & 40.4 & 29.5 & 69.6 & 61.6 & 24.3 \\
\rowcolor{lightblue} Gemini-2.5-Flash & Orus \textsc{(M)} & 50.3 & 56.8 & 39.5 & 25.5 & 68.5 & 59.3 & 23.3 \\
 \rowcolor{lightblue} Gemini-2.5-Pro   & Kore \textsc{(F)} & 47.7 & 58.8 & 39.7 & 21.4 & 69.9 & 60.0 & 23.5 \\
\rowcolor{lightblue} Gemini-2.5-Pro   & Orus \textsc{(M)} & 46.7 & 54.0 & 40.3 & 23.1 & 72.7 & 59.3 & 21.4 \\
\rowcolor{lightgray} GPT-4o-Mini & Coral \textsc{(F)} & 55.9 & 62.1 & 45.5 & 20.0 & 71.2 & 60.5 & 22.4 \\
\rowcolor{lightgray} GPT-4o-Mini  & Ash \textsc{(M)}  & 55.7 & 64.0 & 42.9 & 19.2 & 70.2 & 61.9 & 24.7 \\
\midrule
\multicolumn{2}{l}{\textit{Average}} & 51.2 & 59.2 & 41.4 & 23.1 & \textbf{70.4} & 60.4 & 23.3 \\
\bottomrule
\end{tabular}
\caption{{AST soft accuracy of omni-modal models on  \textit{Confetti} across six TTS configurations. 
}}
\label{tab:overall_results}
\end{table*}

\begin{table*}[t!]
\centering
\scriptsize
\setlength{\tabcolsep}{3pt}
\begin{tabular}{ll ccccccc}
\toprule
\textbf{TTS Model} & \textbf{Voice} &  \textbf{GPT-4o} & \textbf{GPT-1.5} & \textbf{GPT-Mini} &
\textbf{Gemini-2.5} & \textbf{Gemini-3.1} & \textbf{Qwen3} & \textbf{Phi-4} \\
& & \textbf{Realtime} & \textbf{Realtime} & \textbf{Realtime} &
\textbf{Flash-Live} & \textbf{Flash-Live} & \textbf{Omni} & \textbf{MM} \\
\midrule
\rowcolor{lightblue} Gemini-2.5-Flash & Kore \textsc{(F)} & 68.9 & 70.2 & 74.5 & 60.8 & 65.7 & 58.3 & 54.0 \\
\rowcolor{lightblue} Gemini-2.5-Flash & Orus \textsc{(M)} & 70.7 & 71.0 & 65.4 & 57.0 & 60.0 & 61.2 & 53.4 \\
\rowcolor{lightblue} Gemini-2.5-Pro   & Kore \textsc{(F)} & 70.3 & 72.6 & 68.7 & 51.0 & 62.7 & 60.6 & 52.6 \\
\rowcolor{lightblue} Gemini-2.5-Pro   & Orus \textsc{(M)} & 70.1 & 66.1 & 63.3 & 57.0 & 62.8 & 60.3 & 50.6 \\
\rowcolor{lightgray} GPT-4o-Mini & Coral \textsc{(F)} & 69.3 & 76.8 & 69.7 & 58.9 & 62.8 & 61.3 & 54.7 \\
\rowcolor{lightgray} GPT-4o-Mini  & Ash \textsc{(M)}  & 71.4 & 74.6 & 70.2 & 56.6 & 66.7 & 61.0 & 56.1 \\
\midrule
\multicolumn{2}{l}{\textit{Average}} & 70.1 & \textbf{71.9} & 68.6 & 56.9 & 63.4 & 60.4 & 53.6 \\
\bottomrule
\end{tabular}
\caption{F1 score of omni-modal models on \textit{When2Call} across six TTS configurations.}
\label{tab:when2call_results}
\end{table*}

\begin{figure}[t]
    \centering
    \includegraphics[width=0.9\linewidth]{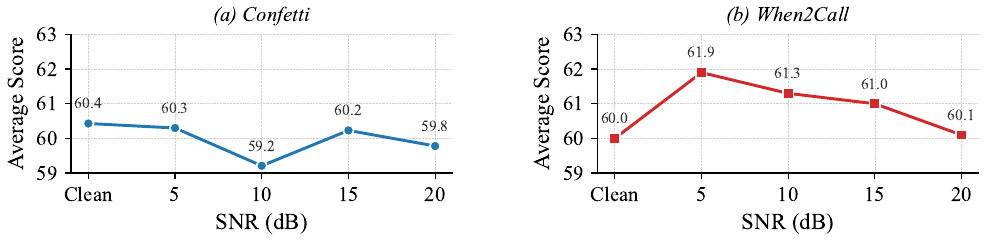}
   
    \caption{\small{Average performance of Qwen3-Omni across SNR levels, aggregated over all TTS models and voices.}}
    \label{fig:snr_comparison}
\end{figure}

\noindent \textbf{Performance across TTS configurations:} Tables \ref{tab:overall_results} and \ref{tab:when2call_results} present tool-calling performance across six TTS configurations (three TTS systems $\times$ two voice types). Performance on Confetti is measured with AST-based soft accuracy, while When2Call is evaluated in terms of F1-score. The results show substantial model- and task-level variation. On Confetti (Table~\ref{tab:overall_results}), Gemini-3.1-Flash-Live obtains the highest average accuracy (70.4\%), followed by Qwen3-Omni (60.4\%), GPT-Realtime-1.5 (59.2\%), GPT-4o-Realtime (51.2\%), GPT-Realtime-Mini (41.4\%), Phi-4-Multimodal (23.3\%), and Gemini-2.5-Flash-Live (23.1\%). On When2Call (Table~\ref{tab:when2call_results}), the relative ranking for the top performing models changes: GPT-Realtime-1.5 leads at 71.9\%, followed by GPT-Realtime-4o (70.1\%) and GPT-Realtime-Mini (68.6\%). Gemini-3.1-Flash-Live, which achieved the best results in Confetti, ranked 4th in When2call (63.4\%), followed by   Qwen3-Omni (60.4\%). Similar to Confetti, Gemini-2.5-Flash-Live and Phi-4-Multimodal performed the worst, achieving F1-scores of 56.9\% and 53.6\%, respectively. The shifts in the overall rankings of different models indicate that audio tool-calling performance cannot be characterized by architecture alone or by evaluation on a single dataset; model choice, dataset, and task formulation all affect outcomes.

\noindent \textbf{Robustness under noisy conditions:} 
We further investigate the robustness to environmental noise. For this, we use the Qwen3-Omni model, which achieves balanced performance across both datasets. Based on the results presented in Figure \ref{fig:snr_comparison}, we observe that its accuracy remains relatively stable across SNR levels, with only modest degradation as noise increases. This result suggests that, under the controlled TTS conditions, moderate environmental noise is not the primary source of performance loss for this model.

\begin{table}[t]
\centering
\scriptsize
\begin{tabular}{lcccccc}
\toprule
\textbf{Model} & \textbf{Clean Text} & \textbf{Direct Voice} & \textbf{Cascade (ASR$\rightarrow$text)}
& \textbf{$\Delta_{\text{TV}}$} & \textbf{$\Delta_{\text{TC}}$} & \textbf{$\Delta_{\text{CV}}$} \\
\midrule
\rowcolor{lightblue}
Gemini-3.1-Flash-Live & \textbf{73.0} & {70.4} & {71.3} & \phantom{0}2.6 & \phantom{0}1.7 & \phantom{0}0.9 \\
\rowcolor{lightgray}
GPT-Realtime-1.5 & 64.0 & 59.2 & 58.8 & \phantom{0}4.8 & \phantom{0}5.2 & -0.4 \\
\rowcolor{lightpurple}
Qwen3-Omni & 62.2 & 60.4 & 58.9 & \phantom{0}1.8 & \phantom{0}3.3 & -1.5 \\
\bottomrule
\end{tabular}
\caption{
\small{Performance on Confetti across clean text-only, direct voice, and ASR$\rightarrow$text cascade settings. 
$\Delta_{\text{TV}}$ measures the text-to-voice gap, 
$\Delta_{\text{TC}}$ measures the text-to-cascade gap, and 
$\Delta_{\text{CV}}$ compares cascade against direct voice 
(Clean Text $-$ Direct Voice, Clean Text $-$ Cascade, and Cascade $-$ Direct Voice, respectively).}
}
\label{tab:confetti_voice_cascade_text}
\end{table}

\subsubsection{Cascade vs.\ Omni Performance Comparison} 
We compare three input settings that correspond to common deployment choices: \textit{clean text}, where the original text input is given directly to the LLM; \textit{direct voice}, where the TTS audio input is given directly to an omni-modal model to measure native speech-based tool calling; and the \textit{ASR-to-text cascade}, where generated audio is first transcribed and then passed to a text-based LLM for tool calling. Overall, clean-text performance estimates the model's tool-calling ceiling, direct voice measures how well that capability transfers to speech input, and the cascade setting tests whether transcription plus text-based reasoning is sufficient. For transcription, we use the GPT-4o-Transcribe\footnote{\url{https://developers.openai.com/api/docs/models/gpt-4o-transcribe}} model. 

Table~\ref{tab:confetti_voice_cascade_text} reports the results on Confetti for the following representative models: Gemini-3.1-Flash-Live, GPT-Realtime-1.5, and Qwen3-Omni-30B-A3B-Instruct. We restrict our model selection to only real-time omni-modal models because voice agents ensure interactive deployment settings by processing speech with low latency and execute function calls during the live conversation. From Table~\ref{tab:confetti_voice_cascade_text}, we find that the text-to-voice gap varies substantially across models: Gemini-3.1-Flash-Live loses 2.6 points from clean text to direct voice (73.0 to 70.4), Qwen3-Omni loses 1.8 points (62.2 to 60.4), and GPT-Realtime-1.5 loses 4.8 points (64.0 to 59.2). This shows that the cost of moving from text to speech is model-dependent rather than a fixed property of omni-modal inference. The cascade results further show that neither architecture uniformly dominates. For Gemini-3.1-Flash-Live, the cascade slightly outperforms direct voice by 0.9 points (71.3 vs.\ 70.4), suggesting that ASR followed by a strong text-mode model can sometimes recover more of the clean-text performance. In contrast, direct voice performs slightly better than the cascade for GPT-Realtime-1.5 by 0.4 points and for Qwen3-Omni by 1.5 points. These differences are modest but important: the better deployment choice depends on the specific model and task, not only on whether the system is implemented as a cascade or an end-to-end omni-modal model. 
Overall, this comparison provides an initial diagnostic signal for teams to evaluate candidate cascade and omni-modal systems on the same underlying tasks, tool schemas, and gold labels, before conducting broader deployment studies involving natural speech, latency, and cost constraints.

\subsubsection{Error Analysis}

\noindent \textbf{Function-call error decomposition on Confetti:}
A practical advantage of our conversion design is that every audio input has a paired text counterpart with the same gold label. This allows us to isolate \emph{paired failure cases}: instances where a model produces the correct tool call in text mode but fails on the corresponding audio input. We analyze these cases on \textit{Confetti} using four mutually exclusive error categories: \emph{i. decision errors}, where the model fails to make the correct high-level tool-use decision; \emph{ii. tool-selection errors}, where it issues a tool call but selects the wrong API; \emph{iii. argument-schema errors}, where it selects the correct tool but omits required arguments, as well as adds invalid arguments; and \emph{iv. argument-value errors}, where it selects the correct tool and argument fields but fills one or more fields with incorrect, incomplete, or misheard content. We pool these paired failures across the six TTS configurations.

Table~\ref{tab:additional-error-analysis} shows that audio-induced failures are not explained by a single error type. For all models, the largest category is argument-value errors, which accounts for 57.2\% failures for  Gemini-3.1-Flash-Live and 54.3\% for GPT-Realtime-1.5. These cases suggest that the models often preserve the broad tool-call structure but fail to recognize the exact argument content from speech. Qwen3-Omni shows a more distributed error profile: argument-value errors remain the largest category at 39.5\%, but argument-schema errors (14.6\%) and tool-selection errors (15.5\%) are more frequent than other models. Decision errors are also substantial across all three systems, ranging from 25.8\% for Gemini-3.1-Flash-Live to 37.4\% for GPT-Realtime-1.5. This suggests that the text-to-audio gap is not only caused by incorrect argument values; in many cases, audio input changes the model's high-level tool-use behavior. Overall, the results indicate that improving speech recognition alone may not fully close the modality gap. Robust audio tool calling also requires preserving the model's downstream decisions about whether to call a tool, which tool to call, and how to call the tool.

\begin{table*}[t]
\centering
\scriptsize
\setlength{\tabcolsep}{3pt}
\begin{tabular}{l|c|c|c|c|c}
\toprule
\textbf{Model} & \textbf{Failure pairs} & \textbf{Decision} & \textbf{Argument value} & \textbf{Argument schema} & \textbf{Tool selection} \\
\midrule
\rowcolor{lightblue} 
Gemini-3.1-Flash-Live & 229 & 25.8\% & 57.2\% & 5.2\% & 11.8\% \\
\rowcolor{lightgray}  
GPT-Realtime-1.5       & 254 & 37.4\% & 54.3\% & 2.8\% & 5.5\% \\
\rowcolor{lightpurple}  
Qwen3-Omni            & 329 & 30.4\% & 39.5\% & 14.6\% & 15.5\% \\
\bottomrule
\end{tabular}
\caption{\small Error decomposition on the \textit{Confetti} dataset. Each row pools the six TTS configurations and includes paired failure cases where the model exactly matches the gold tool call in text mode but fails on the corresponding audio input. \textit{Decision} errors indicate incorrect high-level tool-use behavior; \textit{argument value} errors indicate incorrect or incomplete field values despite the correct tool/field structure; \textit{argument schema} errors indicate missing, invalid, or misassigned arguments; and \textit{tool selection} errors indicate invocation of the wrong API.}
\label{tab:additional-error-analysis}
\end{table*}

\begin{figure*}[t!]
    \centering
    \includegraphics[width=0.95\linewidth]{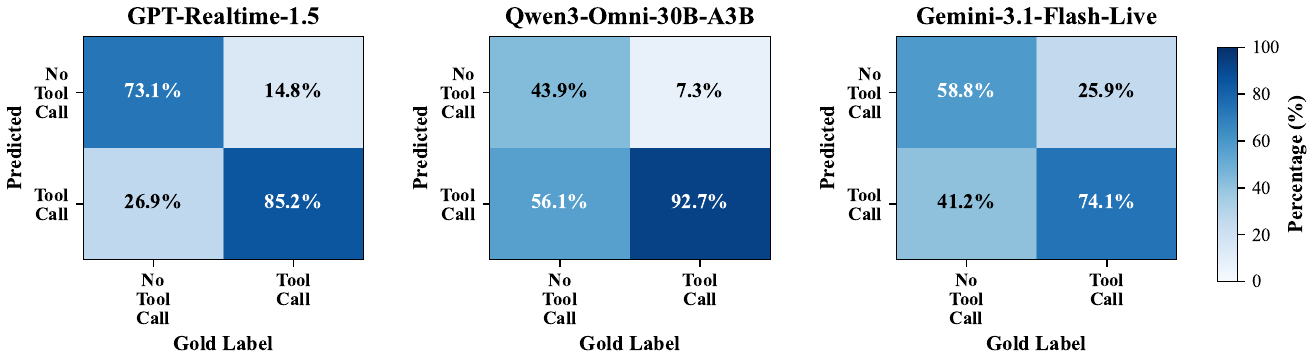}
    \caption{\small Error Analysis on the \textit{When2Call} benchmark computed over 6 TTS voice variants.}
    \label{fig:heatmap_when2call}
\end{figure*}

\noindent \textbf{Tool-call vs.\ no-tool-call errors on When2Call:}
In When2Call, the core challenge is not argument generation but deciding whether a tool call is needed at all. We therefore analyze the confusion matrices in Figure~\ref{fig:heatmap_when2call}, which separate errors on \emph{tool-call} and \emph{no-tool-call} instances. The results show that most models are more reliable when a tool call is required, but struggle in no-tool-call cases.  In particular, Qwen3-Omni correctly identifies tool-call cases in 92.7\% of examples, but correctly identifies no-tool-call cases in only 43.9\%.
GPT-Realtime-1.5 is more balanced, with 85.2\% accuracy on tool-call cases and 73.1\% on no-tool-call cases.
Gemini-3.1-Flash-Live shows a similar asymmetry, performing better on tool-call cases (74.1\%) than no-tool-call (58.8\%) cases. These results suggest that voice-agent failures are not limited to executing tools; models also need 
to know when \emph{not} to call a tool.

\begin{figure}
    \centering
    \includegraphics[width=1\linewidth]{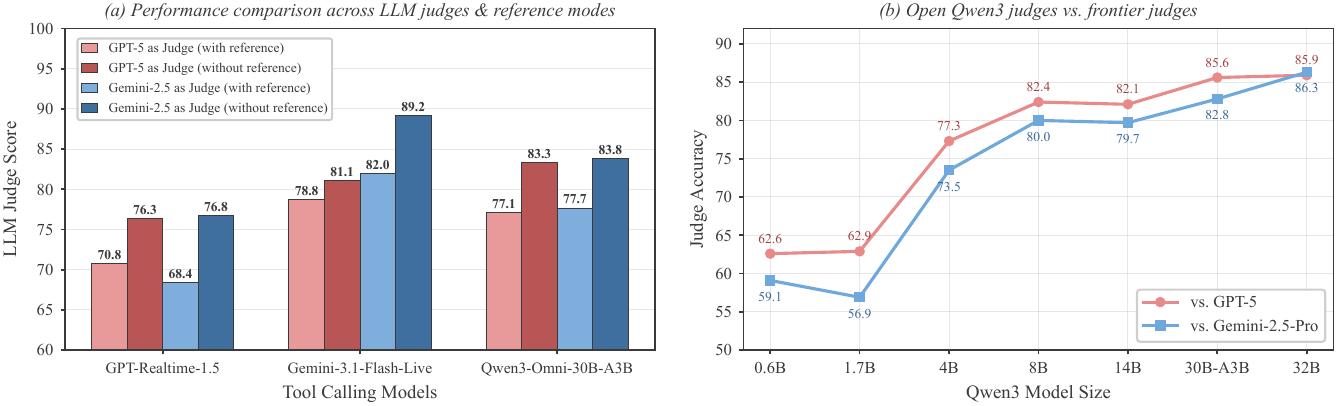}
    \caption{\small{LLM-as-judge evaluation on \textit{Confetti}. 
    (a) Proprietary LLM judge scores in reference-wise and reference-free settings (see Appendix \ref{model_performance_details} for the detailed breakdown). 
    (b) Judgment agreement between open judges (Qwen3) against proprietary reference judgments.}}
    \label{fig:llm_judge}
\end{figure}

\subsubsection{Evaluation using LLM Judges}

Because production deployments often lack complete gold annotations for customer-specific data, we evaluate whether LLM judges can provide a practical reference-free alternative for assessing tool-calling outputs. We focus on the \textit{Confetti} dataset and use GPT-5 and Gemini-2.5-Pro as proprietary LLM judges. We evaluate each output in two settings: a \emph{reference-aware} setting, where the judge sees the gold tool call, and a \emph{reference-free} setting, where the judge needs to assess without seeing the gold tool call labels. 
This comparison tests whether reference-free judging preserves similar model-level conclusions to reference-aware judging (Appendix \ref{llm_judge_prompt} contains the sample prompt for the LLM judge).

\noindent \textbf{Reference-aware vs.\ reference-free judging:}
Figure~\ref{fig:llm_judge}a shows that model rankings are broadly stable across judges and reference modes: Gemini-3.1-Flash-Live receives the highest scores overall, followed by Qwen3-Omni and GPT-Realtime-1.5. Reference-free scores are consistently higher than reference-aware scores for all three models, suggesting that judges may penalize minor deviations from the gold label when references are shown, even if the output is functionally reasonable (see Appendix~\ref{error_analysis_appendix} for an example). To verify this 
  statistically, we apply McNemar's test \citep{mcnemar1947note} to paired judge decisions on the same model
  responses: the reference effect is \textbf{statistically significant} for both judges on
  GPT-Realtime-1.5 and Gemini-3.1-Flash-Live ($p < 0.001$), confirming that the higher reference-free scores reflect a systematic shift in judge decisions rather than sampling noise. We further find that the two judges disagree significantly on
  Gemini-3.1-Flash-Live responses in both settings ($p < 0.001$). Moreover, we observe some
  judge-family specific preference bias: Gemini-2.5-Pro tends to assign higher scores to
  Gemini-3.1-Flash-Live. Overall, reference-free judging provides a useful
  production-oriented signal when gold labels are unavailable. However, results should be
  interpreted with multiple judges rather than a single one.

\noindent \textbf{Human validation of LLM judges:}
To assess whether any specific proprietary judge should be preferred, we sample 200 cases where GPT-5 and Gemini-2.5-Pro disagree and conduct a pairwise preference study with two NLP experts. Gemini-2.5-Pro judgments are preferred in 52\% of cases, while GPT-5 judgments are preferred in 48\%. This near-even split suggests that neither judge is clearly preferred by humans.

\noindent \textbf{Open-source judges for privacy-preserving evaluation:}
To address data privacy and cost concerns associated with proprietary judge APIs, we also benchmark Qwen3 models \citep{yang2025qwen3} of varying sizes as open-source evaluation judges on \textit{Confetti}. Figure~\ref{fig:llm_judge}b shows that Qwen3 judges with at least 8B parameters achieve strong agreement (80\%) with GPT-5 and Gemini-2.5-Pro, with the larger 32B variant exceeding 85\% judgment agreement. This indicates that open judges can provide a viable privacy-preserving evaluation option for organizations that cannot send proprietary 
customer data to the closed judge APIs.

\subsubsection{Text-Only Scaling Analysis}

\begin{figure}[t]
    \centering

    \begin{subfigure}[t]{0.95\linewidth}
        \centering
        \includegraphics[width=1\linewidth]{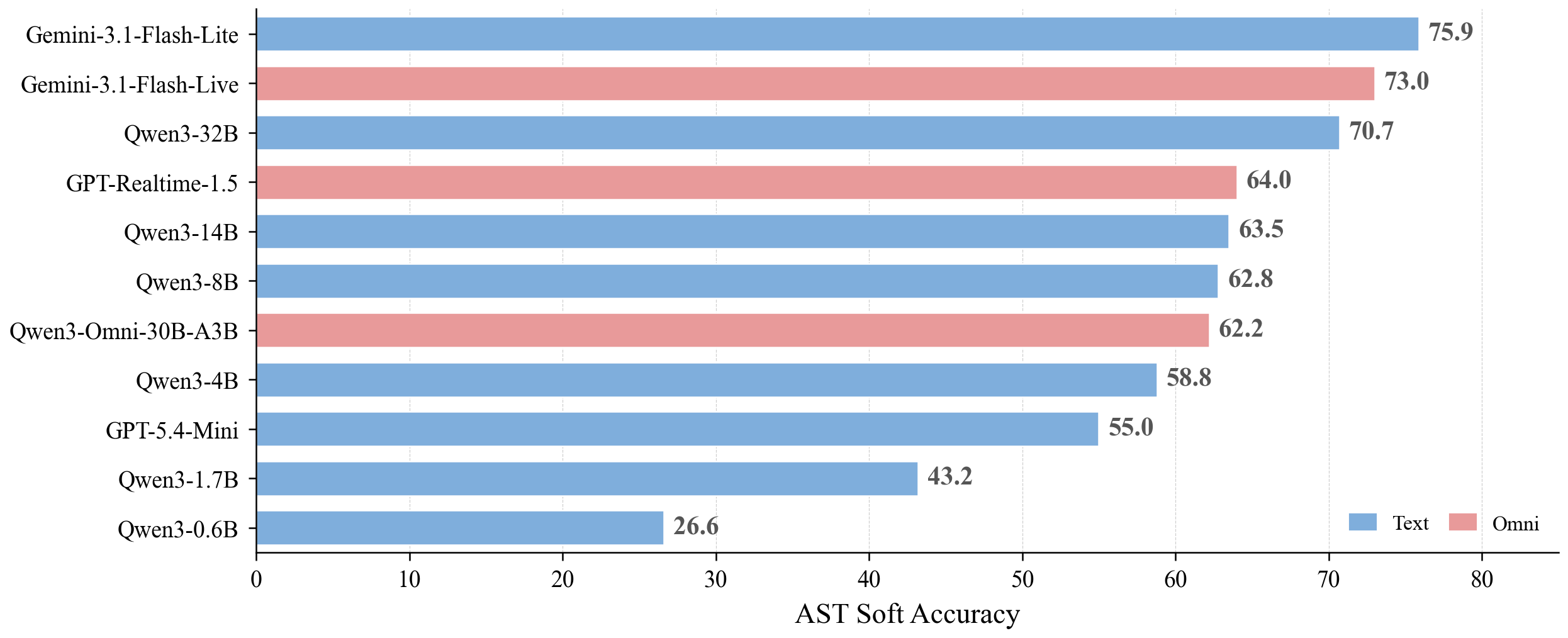}
        \caption{Text-only performance across model families and sizes.}
        \label{fig:text_only_scaling}
    \end{subfigure}

    \vspace{2mm}

    \begin{subfigure}[t]{0.95\linewidth}
        \centering
        \includegraphics[width=0.95\linewidth]{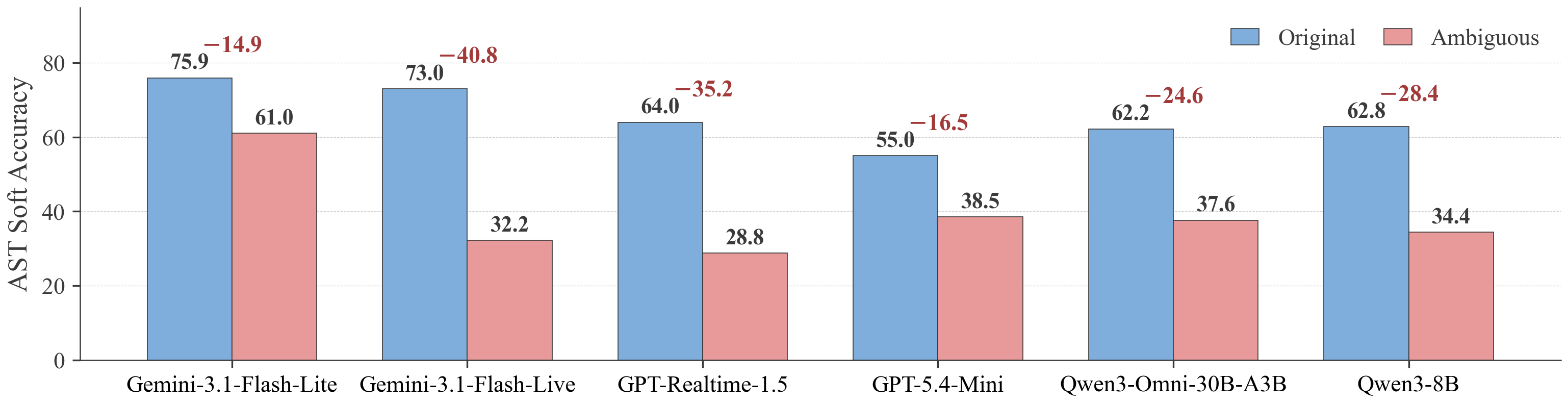}
        \caption{Ambiguous-query reformulation stress test.}
        \label{fig:original_ambigious}
    \end{subfigure}

    \caption{\small Text-only tool-calling analysis on \textit{Confetti}. 
    (a) AST soft accuracy across model families and sizes. 
    (b) AST soft accuracy under ambiguous-query reformulation stress tests.}
    \label{fig:text_scaling_and_robustness}
\end{figure}

The cascade results (Table \ref{tab:confetti_voice_cascade_text}) suggest that text-based tool calling can remain competitive with direct voice models when the audio is first transcribed. This motivates a more deployment-oriented question: if a company uses an ASR$\rightarrow$text-LLM cascade, how strong does the text-only LLM need to be? To answer this, we evaluate text-mode tool-calling performance on \textit{Confetti} across various model families and sizes, including cost-effective Qwen3 models. 
Figure~\ref{fig:text_only_scaling} shows that the accuracy generally increases with model size within the Qwen3 family, rising from 26.6\% for Qwen3-0.6B to 70.7\% for Qwen3-32B. Qwen3-Omni-30B-A3B obtains 62.2\% in text mode, which is comparable to the Qwen3-8B (62.8\%) and Qwen3-14B (63.5\%), suggesting that many smaller text-only models may already provide competitive cascade backends. Among the non-Qwen models, Gemini-3.1-Flash-Lite achieves the highest score (75.9\%), followed by Gemini-3.1-Flash-Live (73.0\%) and GPT-Realtime-1.5 in text mode (64.0\%). 
These results show that clean-text competence varies substantially across model families and sizes. For deployment, this means that audio degradation should be interpreted relative to each model's own text baseline, and that cascades with strong or cost-effective text-only LLMs may be a competitive alternative to direct omni-modal models.

\noindent \textbf{Robustness to query reformulation.}
Beyond the original datasets, we additionally define stress-test slices that practitioners can add when making deployment choices. These slices target customer-support failure modes that may be underrepresented in clean benchmarks and should be treated as a deployment checklist rather than a replacement for the main evaluation. Here, we test whether tool-calling accuracy is robust to ambiguous user phrasing. By introducing vagueness, we generate reformulations of the queries in \textit{Confetti} using GPT-5 and Gemini-2.5-Pro in equal proportion (see Appendix \ref{ambiguous_prompt} for the sample prompt). We then evaluate the performance for the resulting queries in the text-only setting. Figure ~\ref{fig:text_scaling_and_robustness}b shows that ambiguous queries cause large drops across all models. Qwen3-Omni falls from 62.2\% to 37.6\% ($-$24.6 points), while Gemini-3.1-Flash-Lite drops from 75.9\% to 61.0\% ($-$14.9 points). These results suggest that query formulation can be a major source of performance variance even before introducing audio. 

\subsubsection{Practical Implications for Deployment}

Our results suggest that voice-agent architecture should be selected through model- and task-specific evaluation rather than by assuming that cascade or omni-modal systems are uniformly better. Direct omni-modal models are attractive when the text-to-voice gap is small, whereas cascades remain safer when teams need vendor flexibility, or when direct audio introduces large performance degradation. Clean-text performance provides the tool-calling ceiling, and the large drop under ambiguous reformulations shows that handling underspecified requests is as important as improving the audio stack. In practice, teams should use our framework as a first-stage benchmark and choose an omni-modal stack only when it improves accuracy, latency, and cost in the target environment; otherwise, a cascade architecture remains the safer choice.

\section{Conclusion}

We introduced a verifiability-preserving framework for converting text-based tool-calling benchmarks into controlled voice evaluations. By preserving tool schemas and gold labels, the framework enables 
reproducible first-stage comparisons to make 
cascade-vs-omni deployment decisions. Across two benchmarks, three TTS systems, multiple voices, and several LLMs, we find that performance is strongly model- and task-dependent: Gemini-3.1-Flash-Live leads on Confetti, while GPT-Realtime-1.5 leads on When2Call. On Confetti, the text-to-voice gap ranges from 1.8 points for Qwen3-Omni to 4.8 points for GPT-Realtime-1.5, showing that audio degradation is not a fixed property of omni-modal inference.
Moreover, our analysis reveals that paired text--audio conversion provides diagnostic value beyond leaderboard scores. 
Text-only scaling further shows that cost-effective text LLMs can be competitive cascade backends, while ambiguity-based reformulation demonstrates that underspecified user requests can degrade performance even before audio is introduced. Finally, reference-free LLM-as-judge evaluation provides a practical option for production settings without gold labels, and open judges with sufficient scale show strong agreement with proprietary judges.

This study has limitations. We evaluate two datasets and a finite set of current models; broader domains and additional text benchmarks such as BFCL \citep{patil2025bfcl}, $\tau$-bench \citep{yao2025tau,barres2025tau}, as well as MCP benchmarks \citep{wang2025mcp,luo2025mcp,guo2026mcp} remain important extensions. Our audio is TTS-generated, so it should be viewed as a controlled first-stage proxy rather than a replacement for spontaneous calls with natural conversational artifacts. 
Future work should extend the framework to naturally spoken customer-support data \citep{laskar2023ai,balaji2026beyond}, richer stress-test suites, alongside cost--latency evaluation. 


\noindent

\bibliographystyle{tmlr}
\bibliography{custom}

\appendix
\section{Appendix}

\subsection{Ethical Considerations}
We maintain the licensing requirements accordingly while using different tools (OpenAI and Gemini models, HuggingFace, etc.). The human evaluation was conducted by co-authors, and so no additional compensation was required. 

\subsection{Sample Prompts for Omni-Modal LLMs}
\label{omni_llm_prompt}

Our sample prompt in the Confetti dataset is provided below. 

\definecolor{attachedColor}{HTML}{e0efff}
\definecolor{attachedColor2}{HTML}{f3f3f3}
\definecolor{attachedColor3}{HTML}{FFE5CC}
\definecolor{attachedColor4}{HTML}{FFCCCC}
\begin{tcolorbox}[
boxrule=0.25pt,   
  colback=attachedColor2,    
  colframe=black,           
  colbacktitle=attachedColor, 
  coltitle=black,           
  title={{Prompt: Confetti}},
  fonttitle=\bfseries,      
  fontupper=\small          
]

   You are a helpful assistant. \\ 
   
   You will be given a conversation context in text format and an audio input from the user. \\
   
   You are also provided with a list of tools that you can leverage to answer the user query. \\
   
   If a tool is appropriate, return a tool call with clear JSON arguments. \\
   
   If not, answer in plain text. \\

Conversation Context: \{Conversation Context\}\\

List of Tools: \{List of Tools\} \\

Audio Input: \{Audio Query\} 

\end{tcolorbox}
Our sample prompt in the When2Call dataset is provided below. 

\definecolor{attachedColor}{HTML}{e0efff}
\definecolor{attachedColor2}{HTML}{f3f3f3}
\definecolor{attachedColor3}{HTML}{FFE5CC}
\definecolor{attachedColor4}{HTML}{FFCCCC}
\begin{tcolorbox}[
boxrule=0.25pt,   
  colback=attachedColor2,    
  colframe=black,           
  colbacktitle=attachedColor, 
  coltitle=black,           
  title={{Prompt: When2Call}},
  fonttitle=\bfseries,      
  fontupper=\small          
]

   You are a helpful assistant. \\
 
   You will receive an audio input from the user and a list of available tools. \\
 
   Your task is to respond by taking the most appropriate action from the following:\\
 
    1. TOOL\_CALL: If the user's request can be fulfilled by calling one of the available tools, call the tool with appropriate arguments.\\
    
    2. FOLLOW\_UP\_QUESTION: If you need more information to fulfill the request or determine which tool to call, ask a clarifying follow-up question.
     Respond naturally based on the user's audio input. \\
     
    3. CANNOT\_ANSWER: If the user's request 
    cannot be answered based on the information available, just inform the user that you do not know the answer.\\ 
    
    4. DIRECT\_ANSWER: If you know the answer to user's question, just answer it directly.\\ \\



List of Tools: \{List of Tools\} \\

Audio Input: \{Audio Query\} \\

\end{tcolorbox}

\subsection{Sample Prompts for LLM Judge}
\label{llm_judge_prompt}

 \definecolor{attachedColor}{HTML}{e0efff}
\definecolor{attachedColor2}{HTML}{f3f3f3}
\definecolor{attachedColor3}{HTML}{FFE5CC}
\definecolor{attachedColor4}{HTML}{FFCCCC}
\begin{tcolorbox}[
boxrule=0.25pt,   
  colback=attachedColor2,    
  colframe=black,           
  colbacktitle=attachedColor, 
  coltitle=black,           
  title={LLM Judge Prompt (Without Reference): Confetti},
  fonttitle=\bfseries\small,      
  fontupper=\small          
]

You are an expert LLM judge and your goal is to evaluate whether an AI agent generated response in a customer-agent conversation is accurate to address the customer's concern. An AI Agent response can be accurate based on the following two criteria:  \\

(i) It has accurately identified the right tool from the list of available tools to address customer's concern in a given conversation.  \\
(ii) It has accurately identified that none of the tools are relevant to address customer's concern and so it did not make any tool calls. \\

You are first provided with the conversation, a list of available tools, and the AI agent generated response.  You should response with 1 if the AI agent is correct and 0 if the AI agent is incorrect. Do not provide any additional information except `1' or `0'  in your response. \\

The conversation, the list of available tools, and the AI agent generated response are given below.  \\

Conversation: \{Conversation\} \\

Available Tools: \{Tools\} \\

AI Agent Response: 
\{Ai\_Agent\_Response\} \\

\end{tcolorbox}

\subsection{LLM Judge Evaluation Analysis}
\label{error_analysis_appendix}

\paragraph{Example 1: Comparing Judgments in Reference-Wise vs Reference-Free Settings}

\begin{itemize}
  \item \textbf{Conversation:} The user asks for information about the capital of Jamaica.

  \item \textbf{Model Response:} ``The capital of Jamaica is Kingston. It's a vibrant city
  known for its rich culture, music, and history. Here are some key highlights
  about Kingston:...'' (followed by a detailed and accurate description of
  Kingston).

  \item \textbf{Gold Answer:}

  [\{`RAG\_query': \{`query': [`capital of Jamaica']\}\}]

  \item \textbf{Reasoning:} In this case, the model provided a comprehensive and
  factually correct answer directly to the user's question. However, in the reference-wise setting, it was
  judged ``\texttt{incorrect}'' by both LLM judges because the expected action in the \texttt{gold\_answer} was a
  \texttt{RAG\_query} tool call to retrieve the information, rather than
  providing it directly. The model's response was correct and helpful, but it
  did not follow the prescribed tool-use path in the gold label. Nonetheless, in the reference-free setting, it was judged as ``\texttt{correct}'' by the judges. This case explains the reason behind models achieving higher accuracy in the reference-free setting (which we also find in some other instances in the dataset).   
\end{itemize}

\paragraph{Example 2: Comparing Gemini-2.5-Pro Judgments with GPT-5}

\begin{itemize}
  \item \textbf{Conversation:} The user asks to find a hotel in Portland.

  \item \textbf{Model Response:} ``Sure, I can help with that. Could you please provide
  the dates of your stay and any specific preferences you have, like star rating
  or amenities?''

  \item \textbf{Gold Answer:}
  
  [\{`BookHotel\_search\_hotel': \{`location': [`Portland']\}\}]

  \item \textbf{Reasoning:} Here, the user asked to find a hotel, and the model
  responded with a clarifying question to gather more details (dates,
  preferences) before searching. While this is a reasonable conversational step
  to refine the search, the \texttt{gold\_answer} expected an immediate
  \texttt{BookHotel\_search\_hotel} tool call with the available information
  (\texttt{location=[`Portland']}). The model's response was a sensible
  interaction, but it did not execute the specific tool call expected by the
  gold standard at that turn, leading to a ``\texttt{incorrect}'' judgment in the reference-wise setting. However, similar to the prior example, the Gemini-2.5-Pro considers the response as ``\texttt{correct}'' in the reference-free setting. Although in this example, the GPT-5 model considers this as ``\texttt{incorrect}'' in the reference-free setting. Our human annotators prefer Gemini-2.5-Pro judgment in such a case.
\end{itemize}

\subsection{Sample Prompts for Ambiguous Query Generation}
\label{ambiguous_prompt}

 \definecolor{attachedColor}{HTML}{e0efff}
\definecolor{attachedColor2}{HTML}{f3f3f3}
\definecolor{attachedColor3}{HTML}{FFE5CC}
\definecolor{attachedColor4}{HTML}{FFCCCC}
\begin{tcolorbox}[
boxrule=0.25pt,   
  colback=attachedColor2,    
  colframe=black,           
  colbacktitle=attachedColor, 
  coltitle=black,           
  title={LLM Judge Prompt (Without Reference): Confetti},
  fonttitle=\bfseries\small,      
  fontupper=\small          
]

You will receive the conversation context and the original final query asked by a user. You need to rewrite the final user query in an ambiguous form while preserving the user's intent (use the conversation context for grounding). \\

For query reformulation, you can replace specific named entities, dates, numbers, and parameters with pronouns or generic descriptors (e.g., 'San Diego Airport' -> 'this airport', '2025-01-15' -> 'that day'). However, the query should still be answerable when read together with the conversation context. \\

[Conversation Context] \\

[Original Query]

\end{tcolorbox}



\subsection{Additional details in model performance}
\label{model_performance_details}
We provide a detailed breakdown of model performance based on LLM Judge evaluation in Table \ref{tab:confetti-benchmark-combined}.


\begin{table*}[t]
\centering
\scriptsize
\setlength{\tabcolsep}{2.8pt}
\begin{tabular}{ll|ccc|ccc|ccc}
\toprule
\textbf{TTS Model} & \textbf{Voice} &
\multicolumn{3}{c}{\textbf{GPT-Realtime-1.5}} &
\multicolumn{3}{c}{\textbf{Gemini-3.1-Flash-Live}} &
\multicolumn{3}{c}{\textbf{Qwen3-Omni}} \\
\cmidrule(lr){3-5} \cmidrule(lr){6-8} \cmidrule(lr){9-11}
 &  & \textbf{GPT-5} & \textbf{Gemini-2.5-Pro} & \textbf{Avg.}
    & \textbf{GPT-5} & \textbf{Gemini-2.5-Pro} & \textbf{Avg.}
    & \textbf{GPT-5} & \textbf{Gemini-2.5-Pro} & \textbf{Avg.} \\
\midrule
\multicolumn{11}{c}{\textbf{With reference}} \\
\midrule
Gemini-2.5-Flash & Kore  & 71.43 & 69.10 & 70.27 & 77.08 & 79.73 & 78.41 & 78.74 & 78.07 & 78.41 \\
Gemini-2.5-Flash & Orus  & 70.53 & 66.56 & 68.55 & 77.48 & 81.46 & 79.47 & 76.49 & 77.81 & 77.15 \\
Gemini-2.5-Pro   & Kore  & 71.29 & 67.99 & 69.64 & 78.22 & 82.51 & 80.37 & 77.89 & 78.22 & 78.06 \\
Gemini-2.5-Pro   & Orus  & 64.12 & 63.12 & 63.62 & 81.40 & 82.39 & 81.90 & 75.75 & 76.08 & 75.92 \\
GPT-4o-Mini      & Coral & 73.80 & 71.25 & 72.53 & 78.91 & 83.71 & 81.31 & 77.00 & 77.64 & 77.32 \\
GPT-4o-Mini      & Ash   & 73.48 & 72.20 & 72.84 & 79.55 & 82.11 & 80.83 & 76.68 & 78.27 & 77.48 \\
\midrule
\textbf{Average} & -- & \textbf{70.77} & \textbf{68.37} & \textbf{69.57}
& \textbf{78.77} & \textbf{81.98} & \textbf{80.38}
& \textbf{77.09} & \textbf{77.68} & \textbf{77.39} \\
\midrule
\multicolumn{11}{c}{\textbf{Without reference}} \\
\midrule
Gemini-2.5-Flash & Kore  & 76.08 & 79.07 & 77.58 & 81.73 & 87.71 & 84.72 & 83.06 & 84.72 & 83.89 \\
Gemini-2.5-Flash & Orus  & 76.49 & 74.17 & 75.33 & 79.80 & 86.09 & 82.95 & 85.43 & 82.45 & 83.94 \\
Gemini-2.5-Pro   & Kore  & 75.91 & 77.23 & 76.57 & 81.85 & 89.11 & 85.48 & 81.85 & 83.50 & 82.68 \\
Gemini-2.5-Pro   & Orus  & 71.43 & 71.43 & 71.43 & 81.06 & 91.03 & 86.05 & 82.06 & 83.72 & 82.89 \\
GPT-4o-Mini      & Coral & 80.19 & 80.19 & 80.19 & 80.19 & 91.05 & 85.62 & 83.07 & 84.03 & 83.55 \\
GPT-4o-Mini      & Ash   & 77.96 & 78.59 & 78.28 & 81.79 & 90.10 & 85.95 & 84.66 & 84.66 & 84.66 \\
\midrule
\textbf{Average} & -- & \textbf{76.34} & \textbf{76.78} & \textbf{76.56}
& \textbf{81.07} & \textbf{89.18} & \textbf{85.13}
& \textbf{83.35} & \textbf{83.85} & \textbf{83.60} \\
\bottomrule
\end{tabular}
\caption{\small{LLM-judge results on Confetti across TTS configurations for GPT-Realtime-1.5, Gemini-3.1-Flash-Live, and Qwen3-Omni. GPT-5 and Gemini-2.5-Pro are used as judges under reference-aware and reference-free settings. The Avg. columns report the mean of GPT-5 and Gemini-2.5-Pro scores for each model. Higher values indicate better.}}
\label{tab:confetti-benchmark-combined}
\end{table*}

\end{document}